%% file: main.tex
\documentclass[%
 reprint,
 superscriptaddress,
 hyperref,
 amsmath,amssymb,
 aps,
]{revtex4-2}

\input{align_section_title}

\usepackage{hyperref}
\usepackage{color,soul}
\usepackage[dvipsnames]{xcolor}
\definecolor{SLACRed}{RGB}{139,24,27}
\hypersetup{colorlinks,breaklinks,allcolors = SLACRed}

\usepackage[caption=false]{subfig}

\usepackage{booktabs}
\usepackage{graphicx}
\usepackage{dcolumn}
\usepackage{bm}

\input{title}

\begin{document}


\title{\mytitle}

\input{author_affiliation}
\date{\today}

\begin{abstract}
The development of X-ray Free Electron Lasers (XFELs) has opened numerous opportunities to probe atomic structure and ultrafast dynamics of various materials. Single Particle Imaging (SPI) with XFELs enables the investigation of biological particles in their natural physiological states with unparalleled temporal resolution, while circumventing the need for cryogenic conditions or crystallization. However, reconstructing real-space structures from reciprocal-space x-ray diffraction data is highly challenging due to the absence of phase and orientation information, which is further complicated by weak scattering signals and considerable fluctuations in the number of photons per pulse. In this work, we present an end-to-end, self-supervised machine learning approach to recover particle orientations and estimate reciprocal space intensities from diffraction images only. Our method demonstrates great robustness under demanding experimental conditions with significantly enhanced reconstruction capabilities compared with conventional algorithms, and signifies a paradigm shift in SPI as currently practiced at XFELs.
\end{abstract}

\maketitle

The pursuit of detailed understanding of material structures and dynamics at increasingly finer spatial and temporal resolutions continues to drive scientific progress. X-ray Free Electron Lasers (XFELs), featured by their ultrafast and ultrabright x-ray beams, have substantially advanced this endeavor with significant achievements across a wide spectrum of disciplines, such as ultrafast dynamics in condensed matter physics \cite{buzzi2019measuring,mcbride2019phase,plumley2023ultrafast}, planetary science \cite{kraus2017formation,falk2018experimental}, and laboratory astrophysics \cite{takabe2021recent, gaus2021probing}, among other burgeoning fields \cite{le2018fusion, lindroth2019challenges, orville2020recent}.

In particular, XFELs have facilitated novel approaches in observing the structures of biomolecules like viruses, proteins, and nucleic acids through Single Particle Imaging (SPI) \cite{schlichting2012emerging,chapman2019x}. This technique aims to provide single-shot diffraction patterns by directing x-ray photons onto individual isolated objects, which are then captured by area detectors. While cryogenic techniques like x-ray crystallography and electron microscopy (cryo-EM) remain prevalent \cite{shi2014glimpse, bai2015cryo, shoemaker2018x,chua2022better}, they require samples to be frozen, potentially altering or limiting dynamics of biological samples. In contrast, XFEL SPI allows imaging for individual particles at room temperature, maintaining the most natural state of biological samples. Furthermore, the femtosecond resolution of XFEL beams enables not only the elucidation of static structures, but also the observation of dynamic conformational changes, thus offering more comprehensive information. While this technique has achieved significant success in imaging complex virus particles \cite{ekeberg2015three, ekeberg2016single,li2020diffraction}, it faces substantial hurdles including weak signals caused by smaller particles, low photon counts, or a limited number of useful images due to an extremely low hit rate (around $0.1\,\%$ \cite{donatelli2017reconstruction}). These challenges are further magnified by intrinsic complexities of the reconstruction process from highly incomplete information. 

From a theoretical standpoint, reconstructing structures from XFEL SPI data represents a complex inverse problem, extending beyond typical phase retrieval challenges. The intricacy is primarily rooted in the indeterminate orientations of similar particles and the absence of phase information across a vast number of diffraction images, often tens or even hundreds of thousands, necessary for accurate reconstruction \cite{chang2021scaling}. Furthermore, high-resolution reconstruction relies on detecting signals at higher momentum points, where intensity diminishes exponentially. Additionally, experimental conditions introduce further complications, including significant shot-to-shot fluctuation in photon numbers per pulse \cite{li2020diffraction,wang2023specklenn}. These issues collectively complicate the already challenging task of orientation estimation, a critical step for computing the auto-correlation function. These challenges, from both theoretical aspects and experimental conditions, pose pressing needs for innovative approaches in reconstruction algorithms for XFEL SPI. 

\begin{figure*}
    \centering
    \includegraphics[width=0.9\linewidth]{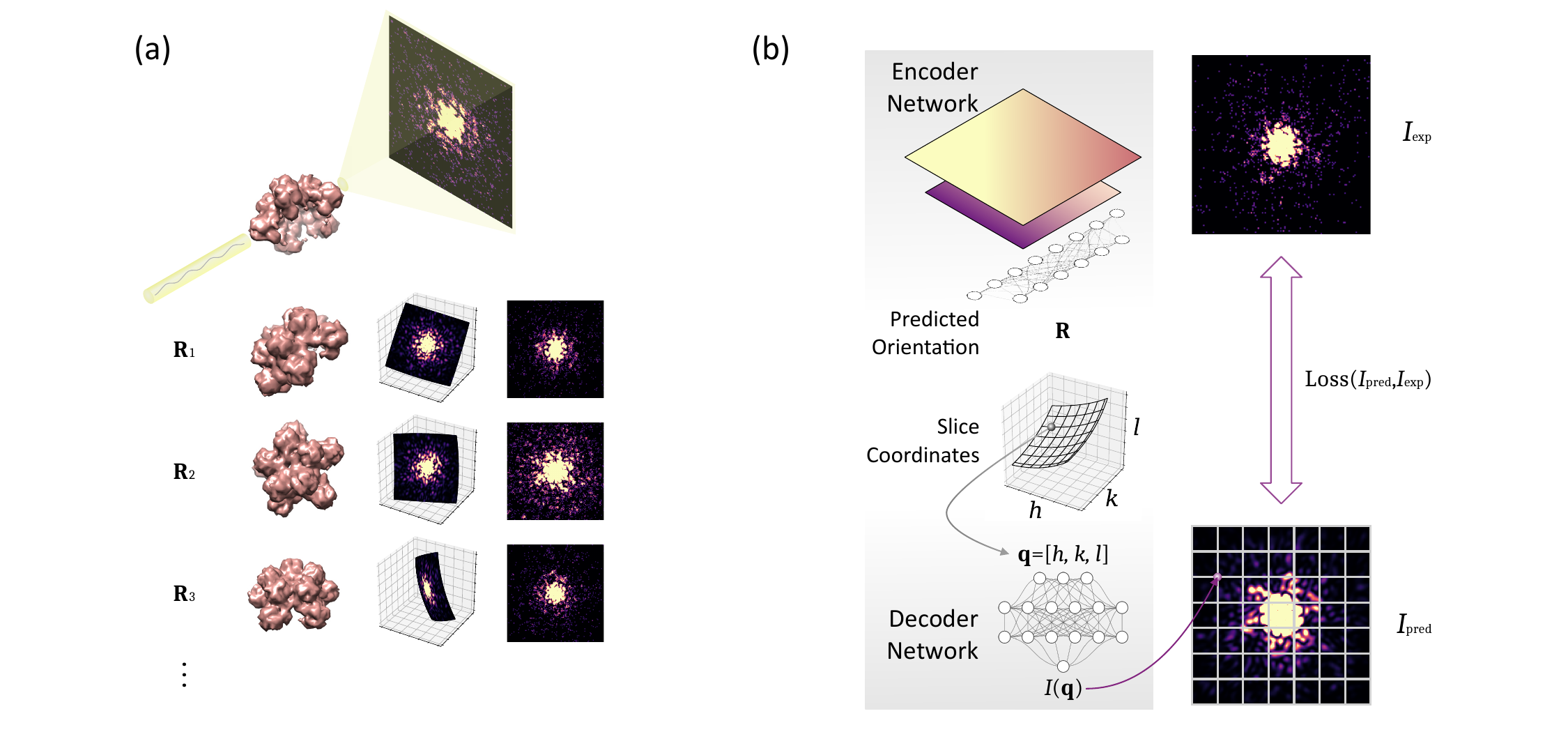}
    \caption{Illustration of Single Particle Imaging (SPI) with X-ray Free Electron Lasers (XFELs) and the proposed machine learning framework. \textbf{(a)} Single-shot diffraction patterns generated by ultafast XFEL pulses and randomly orientated particles; each diffraction pattern corresponds to a slice on the Ewald sphere. \textbf{(b)} The proposed machine learning framework predicts orientations of diffraction patterns by an encoder network and reconstructs the reciprocal space intensity by a decoder network.}
    \label{fig:workflow_illustration}
\end{figure*}

Recent years have witnessed significant advances in ML algorithms for reconstruction from cryo-EM images \cite{Punjani2017,Zhong2021,Gupta2021,Nashed2021,Donnat2022}. Among the novel ML methods, the encoder-decoder architecture, with parameterized $\mathrm{SO}(3)$ orientations as the ``latent codes'', has proven particularly effective. This type of architectures offers straightforward physical insights from the encoder outputs, which enables the decoder to make use of the Fourier slice theorem alongside predicted orientations to predict reconstructed images \cite{Nashed2021,levy2022cryoai}. Moreover, recent development in implicit neural representations, such as the sinusoidal representation network (SIREN) \cite{sitzmann2020implicit}, enables more efficient decoders that yield reconstructions without expensive Fourier transforms over the three-dimensional voxel data, as demonstrated in the CryoAI \cite{levy2022cryoai}. These advancements have not only revolutionized cryo-EM reconstructions, but also provide important inspiration for ML algorithms in reconstructions from XFEL SPI data.

In this work, we present an end-to-end, self-supervised ML algorithm that provides accurate orientation estimations from diffraction patterns and predicts complete reciprocal space intensity predictions, enabling real space reconstructions through phase retrieval. Our method demonstrates exceptional performance and maintains high accuracy and robustness under various challenging experimental artifacts, including significant shot-to-shot photon count fluctuations, strong shot noise and detector read out noise, and missing pixels caused by beam stops. Through comprehensive benchmark tests, we show that this method not only surpasses conventional reconstruction algorithms on challenging datasets, but also can be used in conjunction to improve conventional approaches to deliver reconstructions with even finer resolution. Our method extends the boundary of applications for XFEL SPI, making it possible to investigate structures and conformational dynamics of particles to unprecedentedly smaller scales. Moreover, the demonstrated capability of working with a limited number of images brings promising hope to expedite SPI data acquisition by providing rapid feedback during experiments to refine and streamline data collection strategies.


\section*{Machine learning method for SPI}

In XFEL SPI experiments, x-ray pulses are utilized to generate single-shot diffraction patterns for the same type of, yet randomly oriented, particles. Physically, each diffraction pattern corresponds to reciprocal space intensities $I(\mathbf{q})$ on a collection of momentum points $\mathbf{q}$ that form a slice on the Ewald sphere, as depicted in Figure \ref{fig:workflow_illustration}(a). The intensity $I(\mathbf{q})$ is defined as the squared magnitude of the Fourier transform of electron densities $\rho(\mathbf{r})$ in the real space, namely $I(\mathbf{q})=|\mathcal{F}[\rho(\mathbf{r})]|^{2}$.

The proposed ML-based reconstruction framework consists of two main neural network models: the encoder $\mathcal{E}$ and the decoder $\mathcal{D}$. The encoder takes a detector image as its input and outputs an estimated orientation, described by a rotation matrix in $\mathrm{SO}(3)$, denoted as $\mathbf{R}$. For each image, the predicted rotation matrix is subsequently utilized to map the image pixels to the corresponding reciprocal space coordinates, forming a curved Ewald sphere slice according to the instrument-specific diffraction geometry that is considered known. The decoder serves as an implicit neural representation of the intensity, which predicts the intensity value $I(\mathbf{q})$ given an input momentum point, $\mathbf{q}$. By applying the decoder to all the momentum points in the slice determined from the predicted orientation, a reconstructed diffraction pattern is obtained and further compared with the input image using some chosen loss function to calculate the reconstruction loss. This is then utilized to train the encoder and decoder simultaneously. The overall workflow is schematically displayed in Figure \ref{fig:workflow_illustration}(b), the encoder and decoder evolve towards a self-consistent state during the self-supervised learning process without relying on any external labels.

\begin{figure}
    \centering
    \includegraphics[width=0.9\linewidth]{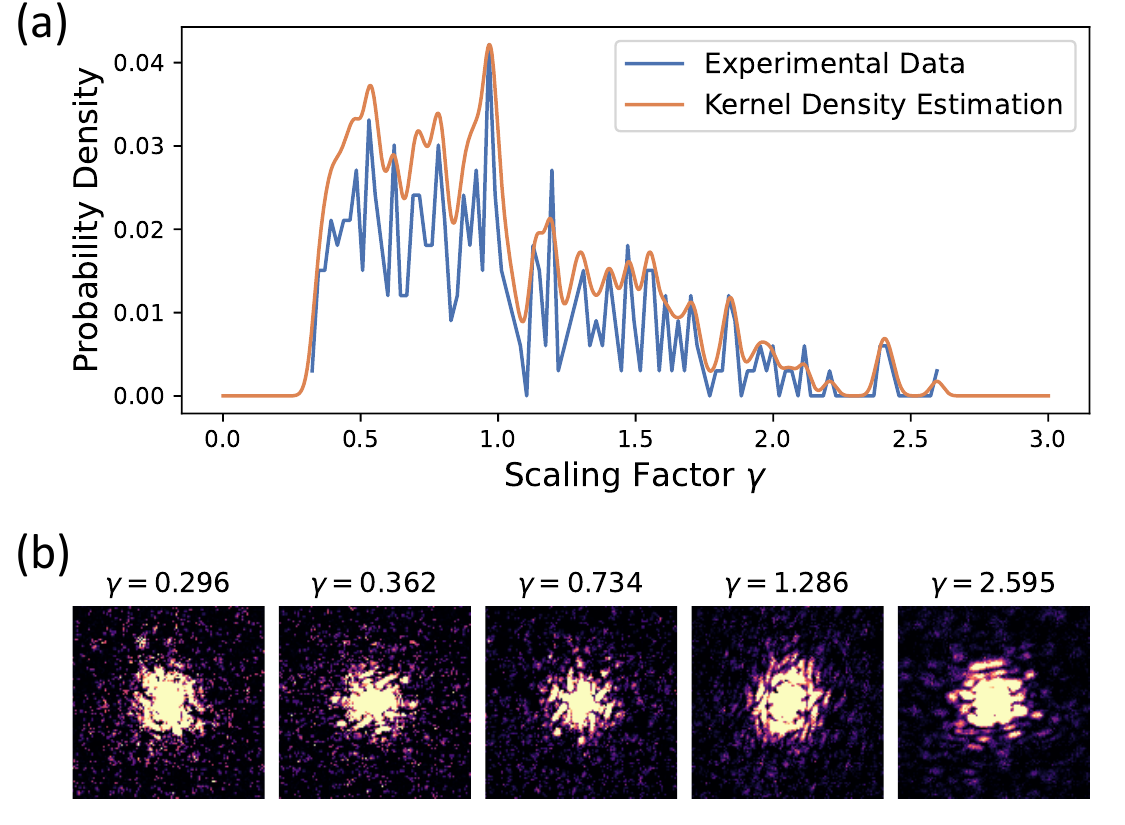}
    \caption{Distribution and effects of the shot-to-shot photon count fluctuations. \textbf{(a)} The experimental and kernel density estimation (KDE)-fitted probability densities. Experimental data is adopted from the curated SPI data of bacteriophage PR772 \cite{li2020diffraction} from Ref.\@ \citenum{wang2023specklenn}. \textbf{(b)} Representative detector images under different levels of photon counts.}
    \label{fig:fluence_jitter}
\end{figure}

\begin{figure*}
    \centering
    \includegraphics[width=0.9\linewidth]{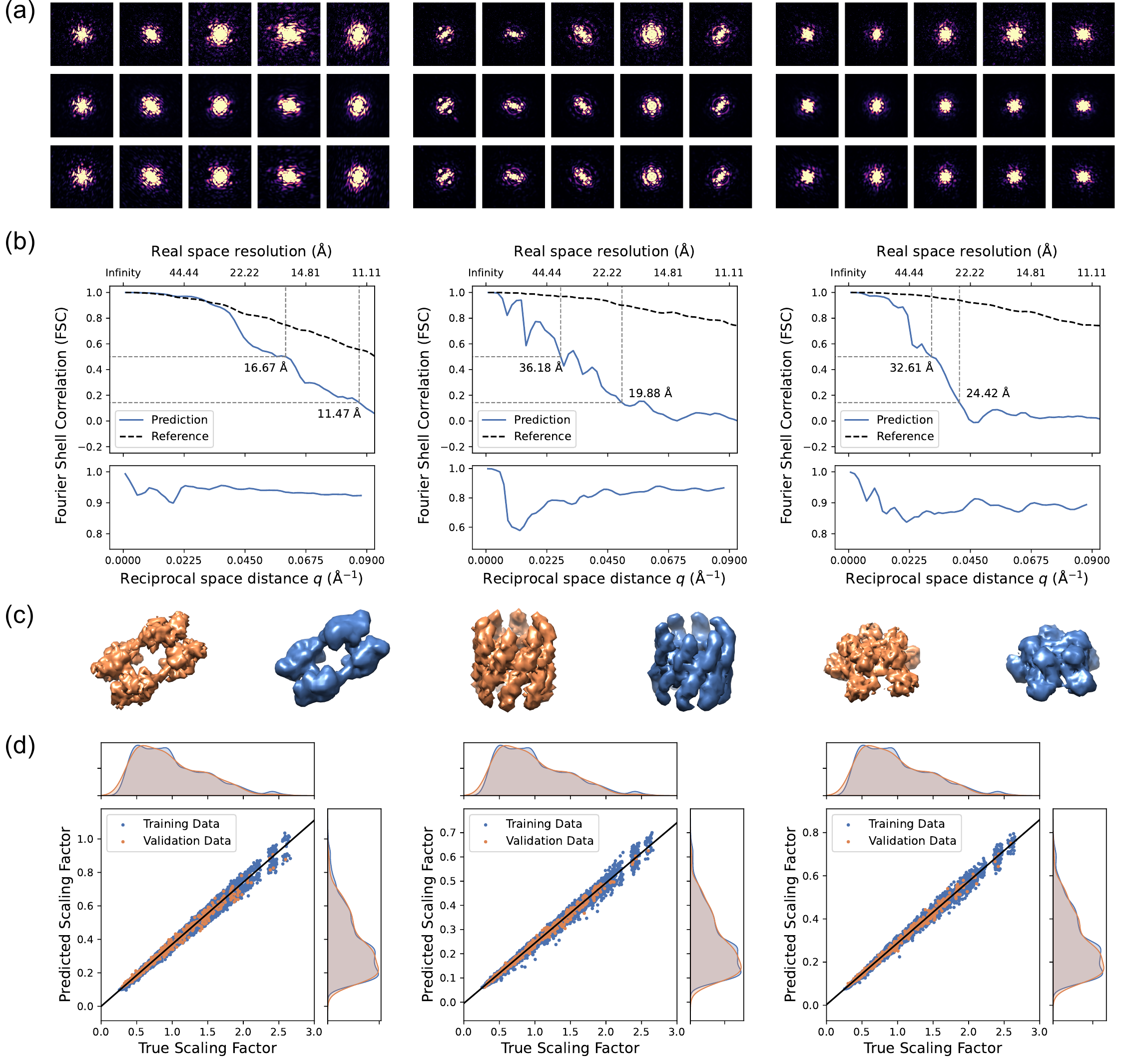}
    \caption{Model performances on detector images with pulse fluctuations, Poisson noise, and detector readout noise across various particle structures. \textbf{(a)} Representative figures of: top) input diffraction patterns; middle) neural network predicted diffraction patterns; bottom) true diffraction patterns. \textbf{(b)} Fourier Shell Correlations (FSC) and estimated resolution of the predicted real and reciprocal volumes. \textbf{(c)} True (left) and reconstructed (right) real space volumes displayed as density isosurfaces. \textbf{(d)} The predicted scaling factors versus true scaling factors.}
    \label{fig:pdb_generalizability}
\end{figure*}

In practice, XFEL SPI measurements are often taken under challenging experimental conditions, particularly with significant fluctuating photon counts per pulse. The fluctuations could lead to highly uncertain photon counts ranging from 0.3 to 2.6 times of the average photon count, as shown by the experimental data in Figure \ref{fig:fluence_jitter}(a). This leads to not only a wide spread of mean intensities across the diffraction patterns, but also to highly different signal-to-noise ratios (SNR) inherent to the Poisson statistics of photon counting. Examples presented in Figure \ref{fig:fluence_jitter}(b) visually illustrate the impacts of these fluctuations, where diffraction patterns are displayed with a upper threshold capped at $0.1\%$ of their maximum values. Diffraction patterns possessing lower number of photons are notably noisier, obscuring physical signals especially at higher momentum points that are critical for detailed reconstructions.

To address the challenges brought by the fluctuating brightness levels, our framework incorporates an auxiliary encoder network, $\mathcal{J}$, that estimates the scaling factor $\gamma$ for each input detector image. The loss function used to train the three neural networks, $\mathcal{E}$, $\mathcal{J}$, and $\mathcal{D}$, is formulated as follows:
\begin{equation*}
    L=\frac{1}{N}\sum_{i=1}^{N}[\ln(\gamma_{\mathrm{pred}} I_{\mathrm{pred}}(\mathbf{q}_{i})+1) - \ln(I_{\mathrm{exp}}(\mathbf{q}_{i})+1)]^{2},
\end{equation*}
where $N$ represents the number of pixels in each detector image. A more comprehensive introduction to the model architecture is presented in \hyperref[sec:model_arch]{Methods}.

\section*{Model performance on diverse structures}

To establish a basic understanding of the model performance across various particle structures, we investigate reconstructions under the experimental condition consisting of photon count fluctuations, Poisson noise, and detector readout noise; details about these experimental artifacts are available in \hyperref[sec:data_prep]{Methods}. We select three protein particles with distinctive geometries to test these conditions, corresponding to Protein Data Bank (PDB) identifiers 1BXR, 3IYF, and 7OK2. Despite the noisy detector images with varying mean intensities, as shown in the top row of Figure \ref{fig:pdb_generalizability}(a), the ML method adeptly reconstructs diffraction patterns (middle row) that closely resemble the true (noise-free) diffraction patterns in the bottom row, suggesting excellent predictions in both orientation and intensity. 

\begin{figure*}[t]
    \centering
    \includegraphics[width=0.9\linewidth]{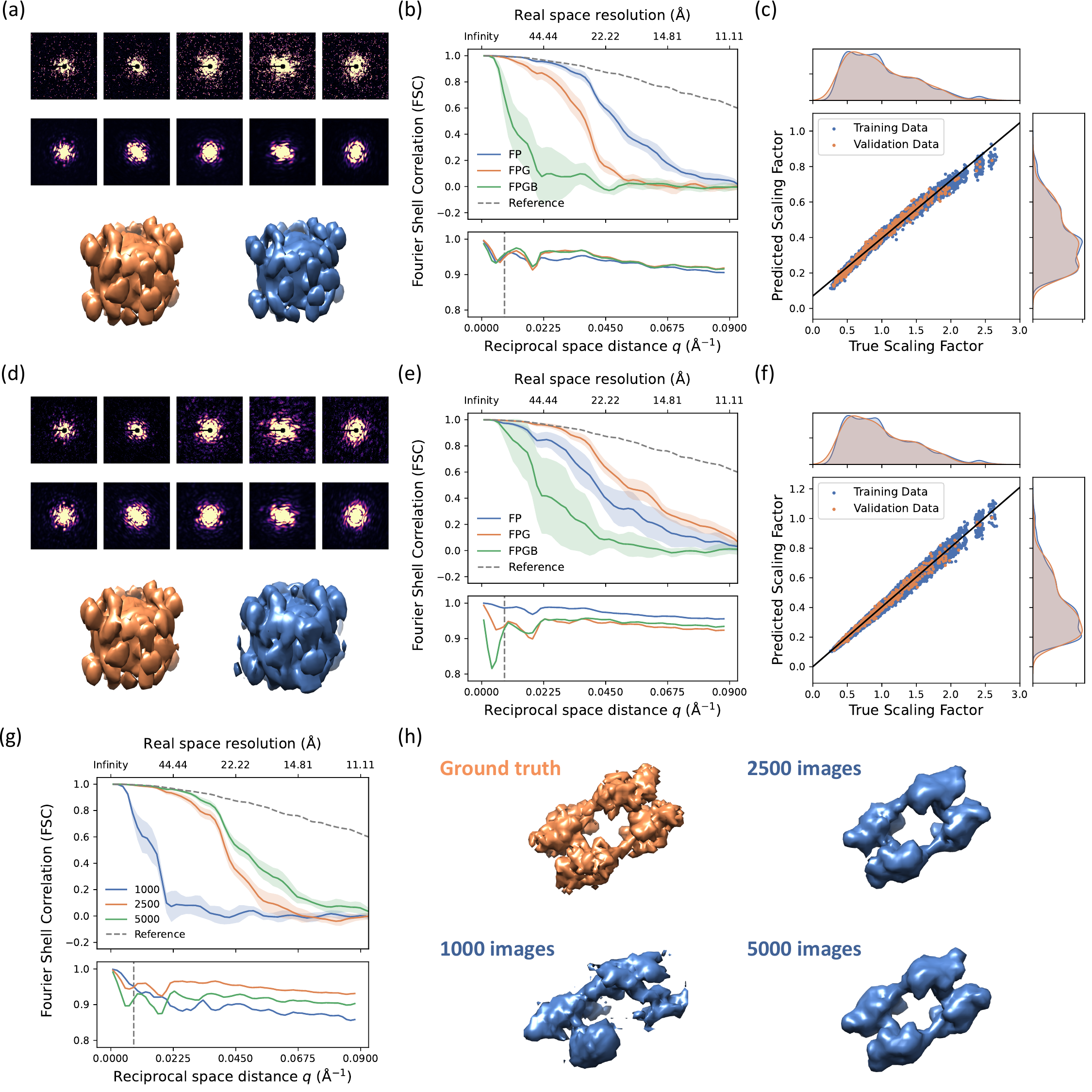}
    \caption{Impacts of various experimental artifacts. \textbf{(a, d)} Representative input (1st row) and predicted (2nd row) diffraction patterns for 1BXR under experimental conditions of photon counts fluctuations, Poisson noise, detector readout noise, and a beam stop mask. Density isosurfaces of the true (left) and predicted (right) intensities are displayed at bottom. \textbf{(b, e)} Fourier Shell Correlations (FSC) and estimated resolutions of the predicted real and reciprocal volumes under various experimental conditions and average beam brightness. The dashed vertical lines in the lower panels indicate the radius, $q_{\mathrm{mask}}\approx 8.26\times10^{-3}\,\mathrm{\AA}^{-1}$, of the center circular part of the beam stop mask. \textbf{(c, f)} The predicted photon counts versus true photon counts under different mean photon numbers (photons/pulse). \textbf{(g)} Summarized FSC metrics of models trained and validated with datasets containing 1000, 2500, and 5000 detector images. \textbf{(h)} Isosurfaces of the true and reconstructed electron densities with the best resolution out of 10 phase retrieval reconstructions. Panels (a-c) correspond to the flux of $10^{13}$ photons/pulse, while panels (d-h) represent results for the flux of $10^{14}$ photons/pulse. Results shown in panels (g) and (h) are obtained with the experimental artifacts FPG.
    }
    \label{fig:experiment_condition}
\end{figure*}

Since the proposed method does not directly predict any real space information, an additional phase retrieval process (denoted as $\mathcal{P}$) is required to further reconstruct the electron density, namely $\rho_{\mathrm{pred}}(\mathbf{r})=\mathcal{P}[I_{\mathrm{pred}}(\mathbf{q})]$. The resolution of the real space reconstructions are estimated using Fourier Shell Correlations (FSC) between $\rho_{\mathrm{pred}}(\mathbf{r})$ and $\rho_{\mathrm{true}}(\mathbf{r})$, at two criteria $0.5$ and $0.143$, as shown in upper panels of Figure \ref{fig:pdb_generalizability}(b). Similarly, we can use FSC to evaluate accuracies of $I_{\mathrm{pred}}(\mathbf{q})$ in the lower panel. More details about FSC calculations are presented in \hyperref[sec:FSC]{Methods}. 

The phase retrieval process is inherently solving a non-trivial inverse problem and introduces a certain level of performance degradation. Therefore, we conduct 10 runs of phase retrieval reconstructions and showcase the best result in Figure \ref{fig:pdb_generalizability}(b) for each particle; the mean and standard deviation of FSCs from all 10 test runs are presented in Section \ref{SI_sec:full_test_pdb_generalizability} of the Supplementary Information. To provide references of how much final resolution is affected by the phase retrieval, we also calculate the FSC between $\rho_{\mathrm{ref}}(\mathbf{r})=\mathcal{P}[I_{\mathrm{true}}(\mathbf{q})]$ and $\rho_{\mathrm{true}}(\mathbf{r})$, as indicated by the dashed curves in Figure \ref{fig:pdb_generalizability}(b). Notably, these reference reconstructions also display considerable decrease in FSCs at higher spatial frequencies, which explains the lower FSCs at high $q=|\mathbf{q}|$ values of real space reconstructions relative to their reciprocal space counterparts. 

The predicted intensities represent one of the direct predictions of our method and display higher FSC with the ground truth. Note that some intricate particle structures pose a challenge to the decoder to learn more complicated intensities in reciprocal space, especially in the range of $q < 0.0225\, \mathrm{\AA}^{-1}$. This is particularly noticeable for the case of 3IYF, where strong variations of the intensity exist in the volume close to the center, as evidenced by their slender diffraction patterns. However, the lowest FSC in intensity for 3IYF still remains above $0.5$, suggesting an overall good agreement with the ground truth. The superior performance of our model on the other two investigated proteins, 1BXR and 7OK2, is likely attributed to their relatively simpler real space structure as well as the corresponding intensities. The intensity prediction for 1BXR is particularly impressive with its FSC being almost consistently above $0.9$. In general, we find that higher intensity FSC leads to better real space resolution, consistent with intuitive expectations. The true and reconstructed electron densities are further visualized with their isosurfaces in Figure \ref{fig:pdb_generalizability}(c), where the reconstructions essentially represent the smoothed counterparts of true volumes without some nuanced details. Remarkably, the predicted volumes recover the overall structure of the true volumes with excellent accuracy.

As for the fluctuation predictor, we note that $I_{\mathrm{exp}}$ is the only available information that the predictions $\gamma_{\mathrm{pred}} I_{\mathrm{pred}}$ are compared with. Since the predicted and true scaling factors are related by $\alpha \gamma_{\mathrm{true}} \times I_{\mathrm{true}} / \alpha$ and consequently $\gamma_{\mathrm{pred}}=\alpha\gamma_{\mathrm{true}}$, any linear relationship passing the origin between $\gamma_{\mathrm{pred}}$ and $\gamma_{\mathrm{true}}$ is considered ideal. In Figure \ref{fig:pdb_generalizability}(d), the fluctuation predictor successfully reveals such linear relationships over the entire range of scaling factors and for various particle geometries; meanwhile, it consistently yields satisfactory predictions on validation images. The increased broadening in the predictions at higher scaling factor values indicates a slight decline in accuracy compared with lower values, likely due to the fewer available training samples with these high photon numbers per pulse. Nevertheless, the overall trends of predictions still align accurately with the fitted lines for all three structures.

\section*{Impact of challenging experimental conditions}

Apart from the three experimental artifacts considered above, we extend our analysis to involve beam stop masks to consider more comprehensive experimental scenarios. The beam stop mask blocks scattered photon from reaching a certain area of the detector, which is replicated in our simulations by applying the binary mask to remove any intensity information from simulated detector images (\hyperref[sec:data_prep]{Methods}). In order to understand how different experimental conditions could affect the reconstruction performance of the proposed ML method, we conducted a series of benchmarks with different combinations of the artifacts, including FP, FPG, and FPGB, with each letter representing a type of experimental artifacts as listed in Table \ref{tab:abbreviations}.

In Figure \ref{fig:experiment_condition}(a) and (d), we showcase some example detector images that include all four experimental artifacts under different mean photon counts, namely $10^{13}$ and $10^{14}$ photons per pulse. Even with considerably affected diffraction patterns, the proposed model still manages to generate predictions that agree well with the input detector images, as evidenced by the middle row of panels (a) and (d). The 3D isosurfaces of the true intensity and two predicted intensities are displayed in the bottom row of the two panels with clear resemblance, demonstrating the impressive robustness of the model under complex experimental conditions.

To investigate how each experimental artifact affects model performance, we summarize the benchmark results in Figure \ref{fig:experiment_condition}(b) and (e) for the two investigated levels of photon counts per pulse. Under lower photon counts, the presence of Gaussian noise and beam stop mask seems to have negligible or even positive impacts on intensity predictions, as hinted by the comparable FSC values. However, it is important to note that FSC mainly provides a more global measure for the agreement between predicted and true volumes and does not necessarily reflect subtleties of detailed reconstruction qualities. This issue is particularly relevant to higher momentum points where the correlations are calculated over larger areas proportional to $q^{2}$. Contrary to similar prediction qualities suggested by FSC evaluated on intensities, we observe that Gaussian noise and beam stop mask lead to increased levels of background noise in their predicted intensities in Figure \ref{SI_fig:intensity_10x} and \ref{SI_fig:intensity_100x} (Supplementary Information Section \ref{SI_sec:intensity_comparison}). These increased noise obscures the already-diminishing physical signals at high spatial frequencies, posing considerable challenges for the phase retrieval to reach better resolutions. Therefore, significant degradation is observed in the FSC evaluated on electron densities when detector readout noise and beam stop mask are added, as shown in the upper panels of Figure \ref{fig:experiment_condition}(b) and (e). An in-depth discussion is provided in Section \ref{SI_sec:intensity_comparison} of Supplementary Information.

For the case with higher photon counts, we find seemingly worse intensity predictions, but better real space reconstructions. This counter-intuitive result is also related to the same insufficiency of detailed information in FSC metrics. In particular, the overall increased intensities in diffraction patterns facilitate the model to capture high frequency details. Thus, even though a similar increase of noise levels is observed in the predicted intensities as more artifacts are added, one can still distinguish between physical signals and noisy backgrounds as illustrated in Figure \ref{SI_fig:intensity_100x} in Supplementary Information. 

Since the beam stop mask consistently covers the center part of every diffraction pattern, any information within the reciprocal space sphere $q<q_{\mathrm{mask}}$ (with $q_{\mathrm{mask}}\approx 8.26\times10^{-3}\,\mathrm{\AA}^{-1}$) is not available for the model to learn. Consequently, we attribute the lower intensity FSC values displayed under the condition FPGB (Figure \ref{fig:experiment_condition}(e)) to the missing information imposed by the beam stop mask. These less accurate intensities combined with elevated levels of noise lead to the most inferior real space reconstruction among three scenarios. Meanwhile, the predicted intensity under the FP condition has overall better intensity predictions than the FPG condition; yet the FPG case gets better electron densities than the FP case. This paradox suggests that the phase retrieval process is not fully optimized and better phase retrieval algorithms are desired, which is further evidenced by failed real space reconstructions on FPGB for 3IYF and 7OK2 in Section \ref{SI_sec:fsc_3iyf_7ok2}; one possible approach is to harness machine learning for phase retrieval as well \cite{wu2021three,yao2022autophasenn}. Since electron densities are not direct outputs of our model, we leave this topic open for future exploration.

The fluctuation predictor displays remarkable robustness against all experimental artifacts with higher photon counts (Figure \ref{fig:experiment_condition}(f)). On the contrary, a non-negligible interception appear under the lower photon counts condition, even though a general linear relationship is still preserved as displayed in Figure \ref{fig:experiment_condition}(c). We further extend the analysis to scenarios with smaller sizes of datasets, comparing to the size of 10000 images for all the rest results presented in this paper. As summarized in Figure \ref{fig:experiment_condition}(g) and (h), the model continues to deliver excellent predictions even with only 2500 detector images available, indicating the promising potential to work with even less useful images that might be caused by, in practice, lower hit rates or limited beamtime allocations. Overall, the excellent performance here highlights the notable resilience of our model under complex experimental conditions.

\begin{table}
    \centering
    \begin{tabular}{ccl}
        \toprule
        Order & Abbreviations & Experimental Artifacts \\
        \midrule
        1 & F & Photon Count Fluctuations \\
        2 & P & Poisson Noise \\
        3 & G & Gaussian Noise \\
        4 & B & Beam Stop Mask \\
        \bottomrule
    \end{tabular}
    \caption{Abbreviations for investigated experimental artifacts. The order represents the sequence of applying different artifacts when multiple ones exist.}
    \label{tab:abbreviations}
\end{table}

\begin{figure*}[t]
    \centering
    \includegraphics[width=0.9\linewidth]{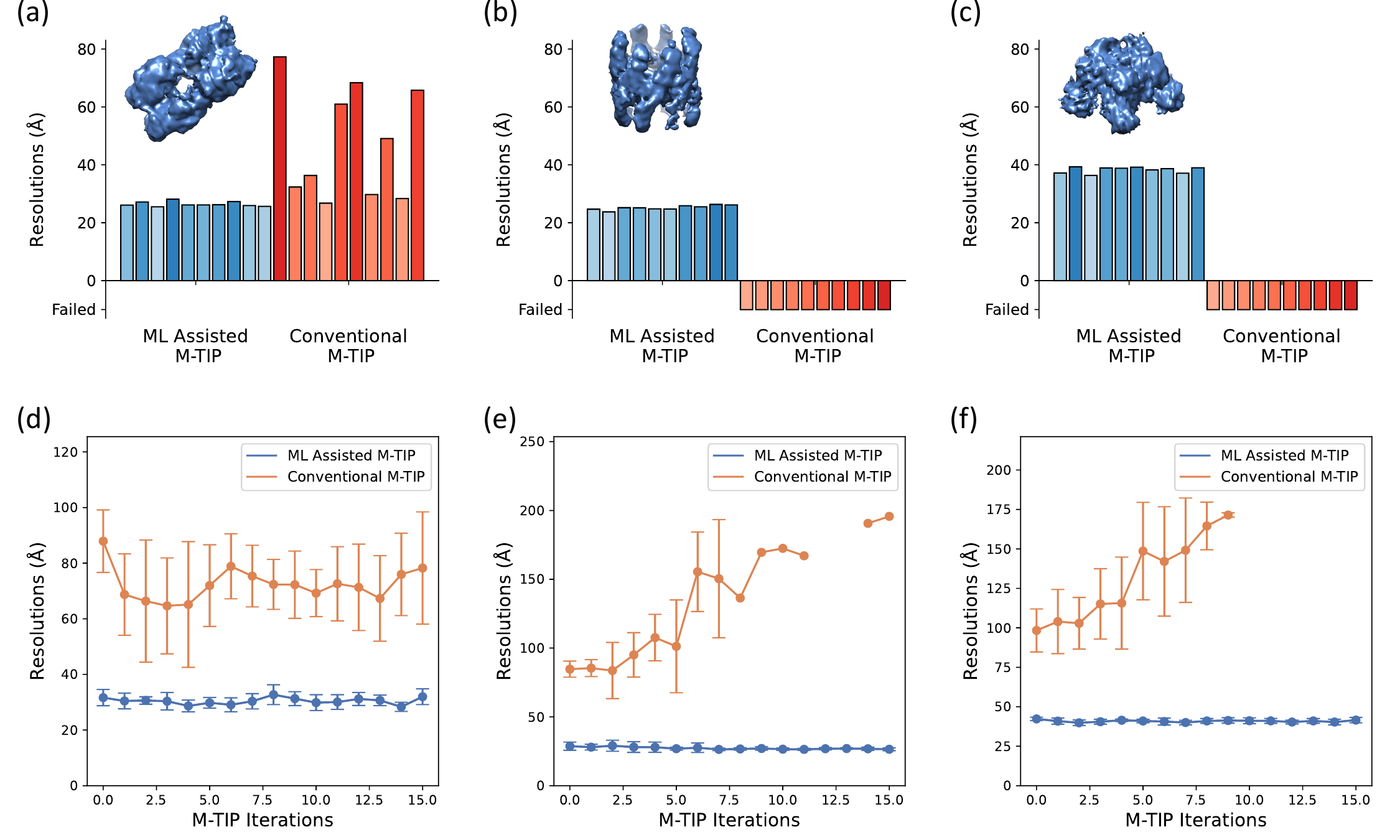}
    \caption{Enhanced performance of conventional M-TIP algorithm with machine learning-predicted orientations and scaling factors. Panels (a-c) display the best real-space resolutions from 10 runs of both the machine learning-assisted and conventional M-TIP algorithms, including inset images of density isosurfaces of the best-resolution reconstructions obtained from the ML-assisted M-TIP. Panels (d-f) show the variation of reconstruction resolution with M-TIP iterations, averaged over 10 runs for both approaches, with error bars representing $\pm1$ standard deviation. Test runs that lead to indeterminate resolutions are considered as failed cases. Three columns of panels represent three structures, 1BXR, 3IYF, and 7OK2, from left to right, respectively.}
    \label{fig:spinifel_benchmark}
\end{figure*}

\section*{Reconstruction from ``unreconstructable'' datasets}

To further investigate how the presented method could facilitate reconstructions from diffraction images taken under challenging experimental conditions, we conduct a series of benchmarks using one of the conventional reconstruction algorithms, Multi-Tiered Iterative Phasing (M-TIP) \cite{donatelli2015iterative,donatelli2017reconstruction,chang2021scaling}. In particular, two reconstruction approaches are tested over the conditions of photon pulse fluctuations, Poisson noise, and detector readout noise: 1) the M-TIP algorithm with predicted orientations for all detector images using the presented ML method; 2) the conventional M-TIP algorithm without any ML-derived priors. Each approach is executed independently 10 times on three protein structures including 1BXR, 3IYF, and 7OK2. For a self-contained discussion, we provide a brief summary of the M-TIP algorithm in \hyperref[sec:m_tip_intro]{Methods}.

Although M-TIP has demonstrated its powerful reconstruction capability on noise-free or moderately noisy datasets, the three experimental artifacts make the already demanding inverse problem even more intractable and often lead to failures of the pure M-TIP approach. As shown in Figure \ref{fig:spinifel_benchmark}(a), while the particularly simpler structure of 1BXR allows the pure M-TIP to achieve some successful reconstructions, the overall performance remains highly unstable with the worst reconstruction reaching a resolution near $80\,\mathrm{\AA}$. Notably, the M-TIP with ML-predicted orientations consistently outperforms the conventional M-TIP with almost every single test reaching the best result from the pure M-TIP approach. As for more complicated structures including 3IYF and 7OK2, the sole application of M-TIP fails to reach any successful reconstructions across all test runs, as depicted by Figure \ref{fig:spinifel_benchmark}(b) and (c). Impressively, every run of the ML-assisted M-TIP approach arrives at a solution with a reasonably good resolution. Furthermore, when comparing with the direct reconstructions using phase retrieval on predicted intensities, i.e., $\rho_{\mathrm{pred}}=\mathcal{P}[I_{\mathrm{pred}}(\mathbf{q})]$, the ML-assisted M-TIP reconstructions reveal much more nuanced details, as evidenced in the inset images of electron density isosurfaces in Figure \ref{fig:spinifel_benchmark}(a-c).

For a closer examination of how ML predictions assist the M-TIP converge, we visualize the iteration-dependent resolutions in Figure \ref{fig:spinifel_benchmark}(d-f). The pure M-TIP approach leads to large mean resolutions and significant variations across all test runs. The reconstruction attempts for 3IYF and 7OK2 are gradually diverging until their FSC metrics fail to reach the criterion $0.5$, leaving their resolutions indeterminate. Meanwhile, the ML-assisted M-TIP approach rapidly reaches convergence in all test runs. The predicted orientations allow the merging step of M-TIP to estimate the auto-correlation function accurately, thereby significantly reducing the otherwise formidable task into a normal phase retrieval problem. The iterative and repeated refinements in M-TIP eventually lead to finer estimations of electron densities. In essence, the integration of the presented ML method with M-TIP not only stabilizes and accelerates the reconstruction process, but also enhances the fidelity of electron density estimations, demonstrating the powerful potential of this hybrid approach in tackling complex datasets where traditional methods falter. This also opens the door to analyze previously unreconstructable datasets.

\section*{Discussion}

In this work, we present an ML method for solving the demanding reconstruction problem in XFEL SPI under challenging experimental conditions. The ML framework includes two encoder networks that predict orientations and relative photon counts from input detector images. This is followed by a decoder that provides an implicit representation of the reciprocal space intensity and can be used for accurate reconstruction of the input images. Training such an ML model only needs the same input information as those required for using conventional reconstruction algorithms, and does not rely on any external data labeling. All three constituent neural networks are trained in a self-supervised way and eventually arrive at a self-consistent state with all detector images reconstructed with a high level of confidence. We demonstrated its generalizability across diverse particle structures and robustness against a wide range of experimental challenges. Remarkably, the ML method displays superior performance compared with the M-TIP algorithm in the investigated scenarios, significantly extending the capability of reconstruction by working in conjunction with the conventional algorithm.

Although the proposed framework has displayed promising potential, some crucial topics still need further investigations. For example, given the lightest particle studied in this paper is 1BXR with a total structure weight of $644.9\, \mathrm{kDa}$, we still have not established a clear understanding on how the model performance will change on smaller particles with more complex structures. Moreover, we have assumed that all the detector images are diffracted from a single protein. However, in practical experiments, diffraction patterns generated from two or more particles exist. Preparing such datasets comprising only single-particle images efficiently within a short period of time for the purpose of real time analysis remains challenging. One potential solution is to adopt recently developed machine learning tools like the SpeckleNN \cite{wang2023specklenn}. In addition, when data compression is applied to experimental data, the detector images might experience the process of compression or decompression, where certain information loss could get introduced, and suggests the need for further studies. Lastly, the non-isotropic structures of various particles might result in non-uniform distribution of orientations with certain preferred poses over $\mathrm{SO}(3)$ after passing through the particle injector, under which scenario the model performance is unknown at this point. 

Of course, further improvements can also be expected by adopting more advanced image encoder models with symmetry-awareness \cite{weiler2019general,cesa2021program,klee2023image} and better decoder models for capturing larger dynamic ranges. Better detectors with higher saturation thresholds might be used to reduce or avoid information loss caused by beam stops. Moreover, machine learning algorithms can be adopted to perform more robust, faster, and more accurate phase retrieval tasks for obtaining better-resolution electron densities from predicted intensities \cite{wu2021three,yao2022autophasenn}.

In summary, this work presents a forward looking study to introduce and investigate an ML framework for highly demanding reconstruction tasks in XFEL SPI. Although this work only investigates simulated detector images due to the lacking available experimental datasets for the studied ranges of particle sizes here, the demonstrated generalizability, robustness, and unprecedented performance comparing with the conventional algorithm promise a wide range of opportunities for researchers to see smaller particles with XFEL. The observed variations of model performances under different experimental conditions further provide crucial insights into optimal experimental conditions, such as detector panel layout or beam stop mask designs. Eventually, this method will empower researchers to navigate through more challenging experimental conditions and to expedite the reconstruction process by utilizing fewer images, even achieving real-time reconstruction during data collection, indicating a revolutionary efficiency and possibility in structural biology and XFEL SPI. Although this work has been mainly focusing on XFEL SPI, such a self-consistent ML framework could also be readily extended into other scattering imaging fields, for example, Time-Resolved Serial Femtosecond Crystallography (TR-SFX) experiments, where challenging scenarios including missing peaks and weak scattering intensities. Moreover, the proposed framework sheds lights on the fundamental inverse problem on real space reconstructions from reciprocal space slices with unknown orientations. This method will serve to enlighten future ML reconstruction algorithms and will eventually facilitate the scientific discovery with SPI as well as other, related scattering and imaging techniques.

\section*{Methods}

\setcounter{figure}{0}
\setcounter{equation}{0}
\setcounter{table}{0}
\setcounter{section}{0}
\renewcommand{\thetable}{M\arabic{table}}
\renewcommand{\thefigure}{M\arabic{figure}}
\renewcommand{\theequation}{M\arabic{equation}}
\renewcommand{\thesection}{M\arabic{section}}

\subsection*{Dataset preparation}
\label{sec:data_prep}

The diffraction patterns are simulated using the Python package \texttt{Skopi} \cite{peck2022skopi} with structures of the three investigated proteins labeled 1BXR, 3IYF, and 7OK2 obtained from the RCSB Protein Data Bank \cite{thoden1999carbamoyl, zhang2010mechanism, ryan2021discovery, berman2000protein, bank_rcsb}. 

The simulations are performed with the Atomic Molecular and Optical (AMO) 86615 beam parameters, including $10^{12}$ photons per pulse with the photon energy $4.6\,\mathrm{keV}$ and a beam radius of $500\,\mathrm{nm}$. A simple square detector geometry is utilized with $128$ pixels spanning $0.1\, \mathrm{m}$ for each side, the distance from detector to sample is set to be $0.2\,\mathrm{m}$. For each protein, a total number of $10000$ noiseless diffraction patterns are simulated using orientations generated randomly based on a uniform distribution over $\mathrm{SO}(3)$ with the random seed $42$. We denote the set of clean images as $\mathcal{S}_{\mathrm{clean}}=\{\mathcal{I}_{1}, \mathcal{I}_{2}, \ldots, \mathcal{I}_{N} \}$ with $\mathcal{I}_{n}\in\mathbb{R}^{128\times 128}$ and $N=10000$, except for examples presented in Figure \ref{fig:experiment_condition}(g) and (h) where $N$ is chosen to be $1000$, $2500$, and $5000$.

For each clean image, $\mathcal{I}_{n}$, we add various experimental artifacts in the following sequence:

\begin{enumerate}
    \item Photon counts fluctuation, given the clean image $\mathcal{I}_{n}$:
    \begin{equation}
        \hspace{1cm}
        \mathcal{I}_{n}^{\mathtt{F}} = \mathtt{F}[\mathcal{I}_{n}^{\ldots}] = \gamma \times \mathcal{I}_{n},
    \end{equation}
    with $\gamma$ being a random variable sampled from the distribution visualized in Figure \ref{fig:fluence_jitter}(a).
    
    \item Poisson noise for photon counting statistics, given the image with any previous artifacts $\mathcal{I}_{n}^{\ldots}$:
    \begin{equation}
    \hspace{1cm}
    \begin{aligned}
        \mathcal{I}_{n}^{\ldots\mathtt{P}} = \mathtt{P}[\mathcal{I}_{n}^{\ldots}] = \{I^{\ldots\mathtt{P}} \mid &  I^{\ldots\mathtt{P}} \sim \mathrm{Poisson}(I^{\ldots}), \\
        & \forall I^{\ldots} \in \mathcal{I}^{\ldots} \}.
    \end{aligned}
    \end{equation}
    
    \item The detector readout noise is modeled by introducing random Gaussian noise with a mean of $\mu=0$ and a standard deviation of $\sigma=0.05$, while keeping the final output image non-negative. Given the image with any previous artifacts $\mathcal{I}_{n}^{\ldots}$:
    \begin{equation}
    \hspace{1cm}
    \begin{aligned}
        \mathcal{I}_{n}^{\ldots\mathtt{G}} = \mathtt{G}[\mathcal{I}_{n}^{\ldots}] = \{ & I^{\ldots\mathtt{G}} \mid  I^{\ldots\mathtt{G}} \sim \max(I^{\ldots} + \sigma, 0), \\
        & \forall I^{\ldots} \in \mathcal{I}^{\ldots} \},
    \end{aligned}
    \end{equation}
    with $\sigma\sim \mathcal{N}(0.0, 0.05)$.

    \item Applying a binary mask to simulate the presence of a beam stop, given the image with any previous artifacts $\mathcal{I}_{n}^{\ldots}$:
    \begin{equation}
        \hspace{1cm}
        \mathcal{I}_{n, ij}^{\ldots\mathtt{B}} = \mathtt{B}[\mathcal{I}_{n}^{\ldots}] = m_{ij} \times \mathcal{I}_{n,ij}^{\ldots},
    \end{equation}
    where $i$ and $j$ represent pixel indices, and $m_{ij}$ is an element for the mask $M$. The binary mask $M$ has $0$'s in the center within the radius of $5$ pixels, the $0$'s extend from the center along the negative $j$-direction till the edge of the detector with the width of $3$ pixels; all the rest area has elements being $1$. A visualization of the complete mask is provided in the Supplementary Information (Section \ref{SI_sec:beam_stop_mask}). 
\end{enumerate}

Therefore, to obtain the detector image with all experimental artifacts, one can apply each artifact in the described order such that $\mathcal{I}_{n}^{\mathtt{FPGB}} = \mathtt{B}[\mathtt{G}[\mathtt{P}[\mathtt{F} [\mathcal{I}_{n}]]]]$.

\subsection*{Model architecture}
\label{sec:model_arch}

The encoder neural network maps a batch of detector images, 
transformed by $\log(1+\mathcal{I})$, into the predicted orientations represented by rotation matrices in $\mathrm{SO}(3)$, namely 
\begin{equation}
    \mathcal{E}: \mathcal{I}\in\mathbb{R}^{128\times 128} \mapsto \mathbf{R}\in\mathbb{R}^{3\times 3}.
\end{equation}
It is based on the residual network (ResNet) with the basic architecture defined by \texttt{resnet18} \cite{he2016deep}. In particular, we made the following changes for the purpose of orientation estimations from detector images:
\begin{itemize}
    \item 
    The weights of the first convolutional kernel are averaged over the input channel dimension such that the original shape $w\in \mathbb{R}^{64\times 3\times 7 \times 7}$ is reshaped into $w\in \mathbb{R}^{64\times 1\times 7 \times 7}$ given that detector images has only one input channel.
    \item 
    The final linear layer is modified such that it outputs 6-dimensional representations of rotations, which is further converted to the corresponding matrix representations \cite{zhou2019continuity, ravi2020pytorch3d}.
\end{itemize}

The fluctuation (fluence jitter) predictor follows a nearly identical architecture as the orientation predictor described above, except for the only difference that outputs of the final linear layer become ReLU (Rectified Linear Unit)-activated scalars,
\begin{equation}
    \mathcal{J}: \mathcal{I}\in\mathbb{R}^{128\times 128} \mapsto \gamma \in\mathbb{R}.
\end{equation}

The decoder neural network maps any reciprocal space coordinate into the predicted scalar that represents a transformed intensity
\begin{equation}
    \mathcal{D}: \mathbf{q}\in \mathbb{R}^{3} \mapsto I(\mathbf{q})\in\mathbb{R}.
\end{equation}
It is adapted from the sinusoidal representation network (SIREN) \cite{sitzmann2020implicit}, in our case, a fully-connected network with sinusoidal activation functions for certain layers, and outputs the transformed intensity such that
\begin{equation}
    y_{\mathrm{pred}}(\mathbf{q}) = \mathrm{SIREN}(\mathbf{q}) + \mathrm{SIREN}(-\mathbf{q}),
\end{equation}
where $y_{\mathrm{pred}}(\mathbf{q}) = \ln(1+I_{\mathrm{pred}}(\mathbf{q}))$, and both the original and its inversed counterparts are passed into the network to enforce the inversion symmetry obeyed by the intensity function due to the real-valued electron density function $\rho(\mathbf{r})$. The SIREN used in this work contains 5 linear layers with sinusoidal activations and an output linear layer with the ReLU activation.

\subsection*{Model training}

All three networks in the model are trained simultaneously using the mean-squared error defined below
\begin{equation}
\begin{aligned}
    L(y_{\mathrm{pred}},\mathbf{R}_{\mathrm{pred}},\gamma_{\mathrm{pred}}; I_{\mathrm{exp}}) = \frac{1}{N}\sum_{i=1}^{N}\bigl[\ln(I_{\mathrm{exp}}(\mathbf{q}_{i})+1) &  \\
    - \ln(\gamma_{\mathrm{pred}} e^{y_{\mathrm{pred}}(\hat{\mathbf{q}}_{i}(\mathbf{R}_{\mathrm{pred}}))}-\gamma_{\mathrm{pred}}+1)& \bigr]^{2}.
\end{aligned}
\end{equation}
Specifically, we firstly extract the intensity from the decoder output by $I_{\mathrm{pred}}=e^{y_{\mathrm{pred}}}-1$ and then multiplying it with the predicted scaling factor $\gamma_{\mathrm{pred}}$. We would like to note that all necessary information to complete training is the intensity data from detector images, i.e., $I_{\mathrm{exp}}$'s. In particular, no orientation label is provided during training, and the loss of orientation predictions, $\mathbf{R}_{\mathrm{pred}}$, are evaluated indirectly through feeding the resulting momentum coordinates $\hat{\mathbf{q}}_{i}(\mathbf{R}_{\mathrm{pred}})$ into the decoder, and then comparing the decoder's outputs with the input images. The $\mathbf{q}_{i}$ in $I_{\mathrm{exp}}(\mathbf{q}_{i})$ is a dummy momentum point for holding the index $i$, and no specific information about $\mathbf{q}_{i}$ is needed.

For the detector images generated with the average of $10^{13}$ photons per pulse, i.e., results in Figure \ref{fig:experiment_condition}(a-c), we multiply the input images by $10$ before passing to the model to obtain similar orders of magnitude in the inputs.

We adopted the Adam optimizer with a weight decay coefficient of $1\times 10^{-4}$ and a cosine learning rate $\alpha$ defined as a function of epoch $x\ge0$ as below
\begin{widetext}
\begin{equation}
    \alpha(x) = 
    \begin{cases}
        \alpha_{\mathrm{\max}} \times \frac{x}{x_{\mathrm{warmup}}}, & 0 \le x \le x_{\mathrm{warmup}}\, ;\\
        \alpha_{\min} + \frac{\alpha_{\max}-\alpha_{\min}}{2}\left[1+\cos\left(\frac{x - x_{\mathrm{warmup}}}{x_{\max} - x_{\mathrm{warmup}}}\right)\right], & x_{\mathrm{warmup}} < x \le x_{\max}\, .
    \end{cases}
\end{equation}
\end{widetext}
In particular, the following parameters are used to calculate learning rates: $\alpha_{\min}=1\times 10^{-7}$, $\alpha_{\max}=3\times10^{-4}$, $x_{\mathrm{warmup}}=5$, and $x_{\max}=1000$.

All models presented in this work are trained with the Distributed Data Parallel (DDP) on four NVIDIA A100 GPUs for a maximum of $1000$ epochs. To ensure reproducibility, a global random seed of $42$ is set. The training dataset contains the first $9500$ diffraction patterns and the rest is set as the validation set. The model weights selected to obtain results in this paper correspond to those reached the lowest validation losses. A few training and validation losses are visualized in Section \ref{SI_sec:loss_hist} of Supplementary Information.

\subsection*{Fourier Shell Correlation (FSC)}
\label{sec:FSC}

The Fourier Shell Correlation (FSC) between two pre-aligned electron densities $\rho_{1}(\mathbf{r})$ and $\rho_{2}(\mathbf{r})$ is defined as
\begin{equation}\label{eq:fsc_definition}
    \mathrm{FSC}(q_{n}) = \frac{\sum\limits_{q\in\mathrm{shell}_{n}} \mathcal{F}[\rho_{1}](\mathbf{q}) \cdot \mathcal{F}[\rho_{2}]^{\ast}(\mathbf{q})}{\left[\sum\limits_{q\in\mathrm{shell}_{n}}\left|\mathcal{F}[\rho_{1}](\mathbf{q})\right|^{2} \cdot \sum\limits_{q\in\mathrm{shell}_{n}}\left|\mathcal{F}[\rho_{2}](\mathbf{q})\right|^{2}\right]^{1/2}},
\end{equation}
where the $n$-th shell is a set of momentum points such that
\begin{equation*}
    \mathrm{shell}_{n} = \left\{\mathbf{q}\, \middle|\, q_{n}-\frac{\Delta q}{2} \le q \le q_{n}+\frac{\Delta q}{2}\right\}.
\end{equation*}
For calculating the FSC between two non-negative intensities $I_{1}(\mathbf{q})$ and $I_{2}(\mathbf{q})$, we replace the $\mathcal{F}[\rho_{i}](\mathbf{q})$ in Eq.\@ \eqref{eq:fsc_definition} with $\sqrt{I_{i}(\mathbf{q})}$ for $i=\{1,2\}$.

\subsection*{Multi-tiered iterative phasing (M-TIP)}
\label{sec:m_tip_intro}

For providing necessary contexts of some hyper-parameters used in this paper and for completeness, we provide a very brief summary for the multi-tiered iterative phasing (M-TIP) algorithm; more detailed introduction to M-TIP is available in the Refs.\@ \citenum{chang2021scaling}, \citenum{donatelli2015iterative}, and \citenum{donatelli2017reconstruction}. In particular, the Python implementation \texttt{SpiniFEL} is adopted for obtaining related results \cite{chang2021scaling, spinifel_repository}.

In general, the M-TIP is an iterative algorithm with each iteration comprising four major steps; suppose that we have the set of detector images, $\mathcal{S}^{\mathrm{exp}}$, and we are at the $j$-th iteration:
\begin{enumerate}
    \item \textbf{Slicing.} Given the current intensity estimation, $I^{j}(\mathbf{q})$, and the set of reference orientations $\mathcal{R}^{\mathrm{ref}}=\{\mathbf{R}_{1}^{\mathrm{ref}},\ldots, \mathbf{R}_{N_{\mathrm{ref}}}^{\mathrm{ref}}\}$ (a grid over $\mathrm{SO}(3)$), the slicing step outputs a set of slices, $\mathcal{S}^{j}=\{\mathcal{I}_{1}^{j},\ldots,\mathcal{I}_{N_{\mathrm{ref}}}^{j}\}$, by performing non-uniform fast Fourier transform (NUFFT) on the auto-correlation function $A^{j}(\mathbf{r})=|\mathcal{F}^{-1}[I^{j}(\mathbf{q})]|$ and calculating the momentum points that correspond to $\mathcal{R}^{\mathrm{ref}}$.
    
    \item \textbf{Orientation Matching.} Given the set of slices, $\mathcal{S}^{\mathrm{exp}}$, this step calculates the pairwise $L_{2}$-distances between the slices sets $\mathcal{S}^{\mathrm{exp}}$ and $\mathcal{S}^{j}$. It then assigns each detector image $\mathcal{I}_{n}^{\mathrm{exp}}$ the orientation from $\mathcal{R}^{\mathrm{ref}}$ with lowest distance as the estimates orientation, such that 
    \begin{equation*}
        \hspace{1cm}
        \mathcal{R}^{j}=\{\mathbf{R}_{m_{1}}^{\mathrm{ref}},\ldots, \mathbf{R}_{m_{N}}^{\mathrm{ref}}\},
    \end{equation*}
    with $m_{n} = \arg\min_{m} L_{2}(\mathcal{I}_{n}^{\mathrm{exp}},\mathcal{I}_{m}^{j})$.
    
    \item \textbf{Merging.} Given the estimated orientations $\mathcal{R}^{j}$ and the set of detector images $\mathcal{S}^{\mathrm{exp}}$, the following optimization problem is solved in the merging step to obtain the desired (flattened) auto-correlation function $A^{j+1}(\mathbf{r})$:
    \begin{equation}
        \hspace{1cm}
        \min_{A} \left| \hat{\mathcal{F}}^{j} A^{j+1} - \mathcal{I}^{\mathrm{exp}} \right|,
    \end{equation}
    where $\mathcal{I}^{\mathrm{exp}}$ is the flattened and concatenated intensity data from $\mathcal{S}^{\mathrm{exp}}$, and $\hat{\mathcal{F}}^{j}$ represents the NUFFT operator that is dependent on $\mathcal{R}^{j}$. 
    
    \item \textbf{Phase Retrieval.} A combination of normal phase retrieval algorithms including the error reduction (ER) and hybrid input-output (HIO) is used to estimate the electron density $\rho^{j+1}(\mathbf{r})$ from the intensity $I^{j+1}(\mathbf{q})$, where the intensity is obtained by $I^{j+1}(\mathbf{q}) = \left|\mathcal{F}[\mathcal{A}^{j+1}(\mathbf{r})]\right|$.
\end{enumerate}

The results presented in Figure \ref{fig:spinifel_benchmark} are obtained with the size of reference orientation set being $20000$. In particular, all orientations in the $\mathcal{R}^{\mathrm{ref}}$ used in the pure M-TIP tests is generated uniformly over $\mathrm{SO}(3)$. The reference orientations used in the ML-assisted M-TIP tests contains $10000$ uniformly generated orientations over $\mathrm{SO}(3)$ and another $10000$ orientations predicted by the proposed model for $\mathcal{S}^{\mathrm{exp}}$. The detector images used for the ML-assisted M-TIP tests are also scaled by the predicted scaling factor through $\mathcal{I}_{n}^{\mathrm{exp}} / \gamma_{n}$. The rest algorithm settings are identical.

\section*{Code Availability}

The source code required to reproduce all results presented in this study will be available at the code repository \href{https://github.com/zhantaochen/neurorient}{https://github.com/zhantaochen/neurorient} with the release version \texttt{v0.0.0} upon submission.

\section*{Acknowledgement}

Z.C.\@ and J.J.T.\@ were supported by the Department of Energy, Laboratory Directed Research and Development (LDRD) program at SLAC National Accelerator Laboratory, under Contract No.\@ DE-AC02-76SF00515. Portions of this work were supported by the U.S.\@ Department of Energy, Office of Science, Basic Energy Sciences under Award No.\@ DE-SC0022216. This research used computational resources of the National Energy Research Scientific Computing Center (NERSC), a U.S.\@ Department of Energy Office of Science User Facility located at Lawrence Berkeley National Laboratory, operated under Contract No.\@ DE-AC02-05CH11231. We acknowledge the assistance of the large language model ChatGPT by OpenAI in refining the language and enhancing the readability of this paper.

\bibliography{references}

\clearpage
\onecolumngrid 
\begin{center}
    \large\textbf{\mytitle} \\ \vspace{10pt} \textbf{Supplementary Information}
\end{center}
\input{supplementary_information}

\end{document}

%% file: align_section_title.tex
\usepackage{etoolbox}
\patchcmd{\section}
  {\centering}
  {\raggedright}
  {}
  {}
\patchcmd{\subsection}
  {\centering}
  {\raggedright}
  {}
  {}

%% file: title.tex
\newcommand{\mytitle}{Augmenting x-ray single particle imaging reconstruction with self-supervised \\ machine learning}

%% file: author_affiliation.tex
\author{Zhantao Chen}
\thanks{Z.\@ Chen and C.\@ Wang contributed equally to this work}
\email[\\]{zhantao@stanford.edu}
\affiliation{%
 Linac Coherent Light Source, SLAC National Accelerator Laboratory, Menlo Park, CA, USA.
}%
\affiliation{%
 Stanford Institute for Materials and Energy Sciences, Stanford University, Stanford, CA, USA.
}%

\author{Cong Wang}
\thanks{Z.\@ Chen and C.\@ Wang contributed equally to this work}
\affiliation{%
 Linac Coherent Light Source, SLAC National Accelerator Laboratory, Menlo Park, CA, USA.
}%

\author{Mingye Gao}
\affiliation{%
 Department of Electrical Engineering and Computer Science, Massachusetts Institute of Technology, Cambridge, MA, USA.
}%

\author{Chun Hong Yoon}
\affiliation{%
 Linac Coherent Light Source, SLAC National Accelerator Laboratory, Menlo Park, CA, USA.
}%

\author{Jana B.\@ Thayer}
\email[]{jana@slac.stanford.edu}
\affiliation{%
 Linac Coherent Light Source, SLAC National Accelerator Laboratory, Menlo Park, CA, USA.
}%

\author{Joshua J.\@ Turner}
\email[]{joshuat@slac.stanford.edu}
\affiliation{%
 Linac Coherent Light Source, SLAC National Accelerator Laboratory, Menlo Park, CA, USA.
}%
\affiliation{%
 Stanford Institute for Materials and Energy Sciences, Stanford University, Stanford, CA, USA.
}%


%% file: supplementary_information.tex
\setcounter{figure}{0}
\setcounter{equation}{0}
\setcounter{table}{0}
\setcounter{section}{0}
\renewcommand{\thetable}{S\arabic{table}}
\renewcommand{\thefigure}{S\arabic{figure}}
\renewcommand{\theequation}{S\arabic{equation}}
\renewcommand{\thesection}{S\arabic{section}}

\section{Detailed comparisons on intensity predictions}
\label{SI_sec:intensity_comparison}

\begin{figure*}[ht]
    \centering
    \subfloat[Photons per pulse $10^{13}$ under the condition \texttt{FP}]{
        \includegraphics[width=0.6\textwidth]{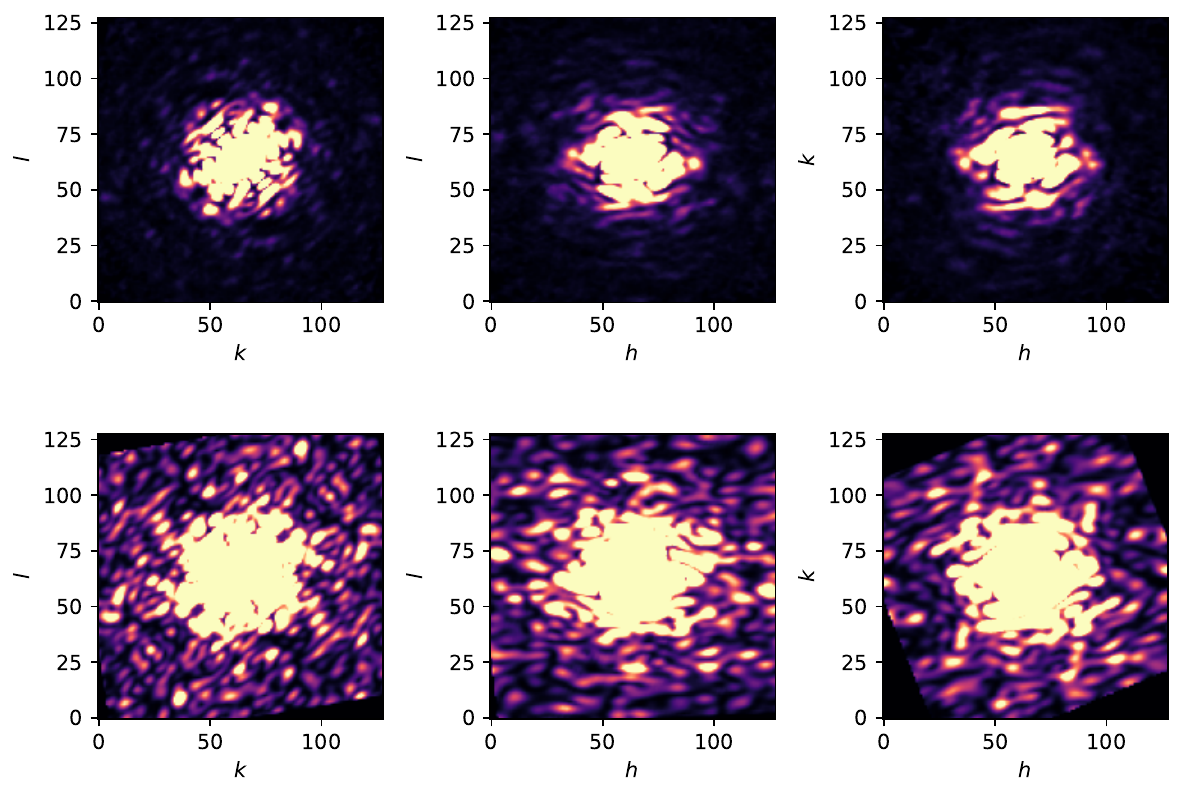}
        \label{SI_fig:fp_10x}
    } \\
    \subfloat[Photons per pulse $10^{13}$ under the condition \texttt{FPG}]{
        \includegraphics[width=0.6\textwidth]{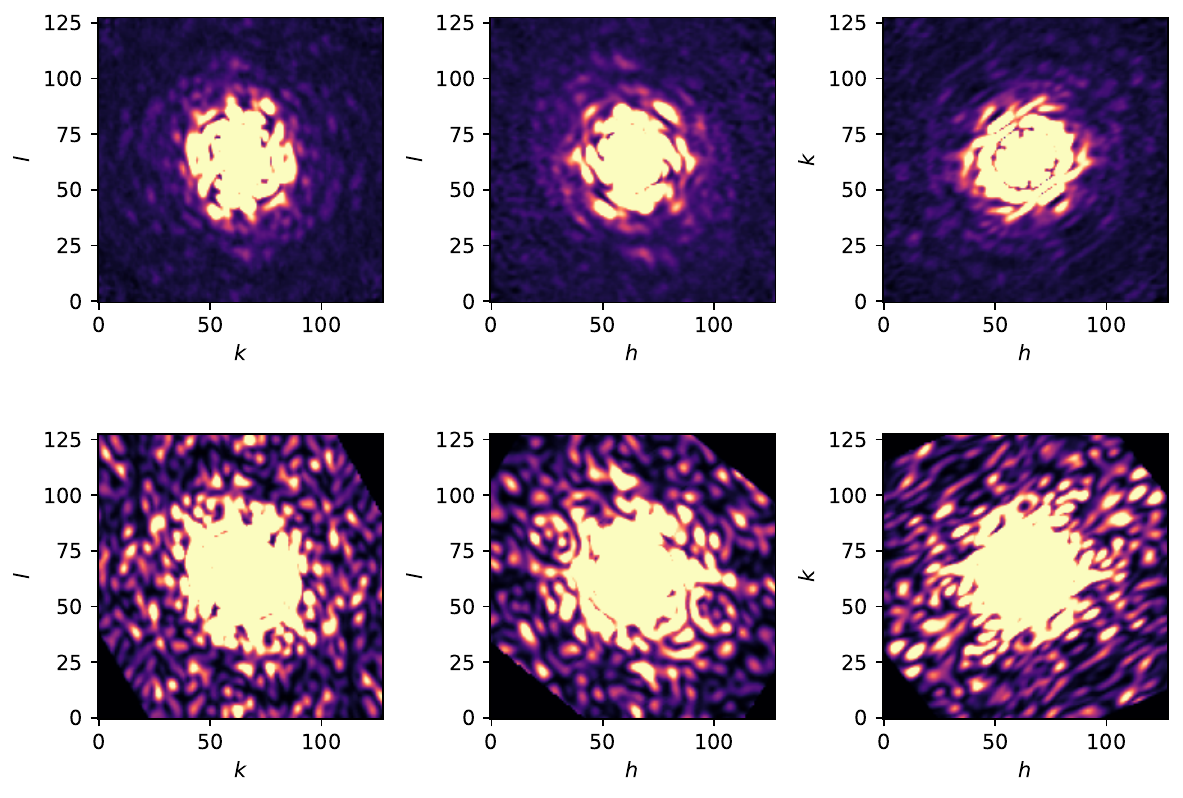}
        \label{SI_fig:fpg_10x}
    } \\
    \subfloat[Photons per pulse $10^{13}$ under the condition \texttt{FPGB}]{
        \includegraphics[width=0.6\textwidth]{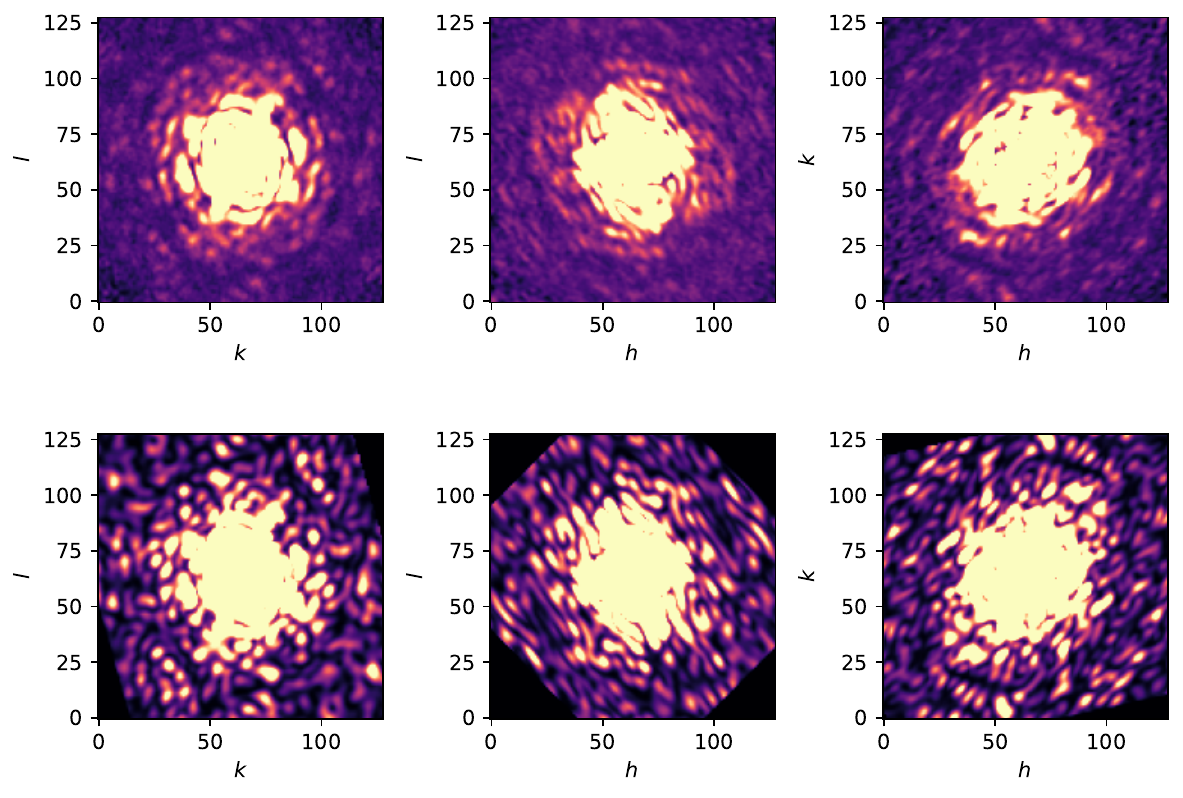}
        \label{SI_fig:fpgb_10x}
    }
    \caption{Predicted (upper panels) and true intensities (lower panels) visualized through cross-sections for models trained with an average of $10^{13}$ photons per pulse under various experimental conditions.}
    \label{SI_fig:intensity_10x}
\end{figure*}

\begin{figure*}[ht]
    \centering
    \subfloat[Photons per pulse $10^{14}$ under the condition \texttt{FP}]{
        \includegraphics[width=0.6\textwidth]{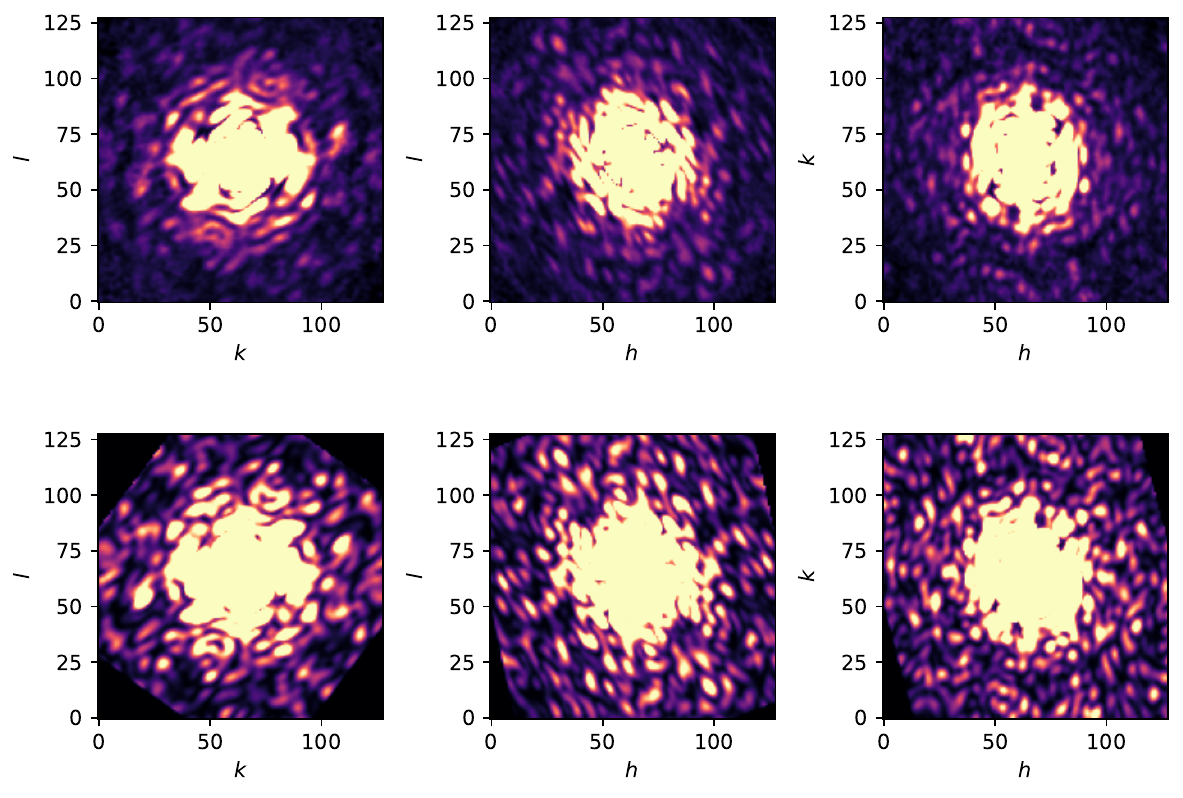}
        \label{SI_fig:fp_100x}
    } \\
    \subfloat[Photons per pulse $10^{14}$ under the condition \texttt{FPG}]{
        \includegraphics[width=0.6\textwidth]{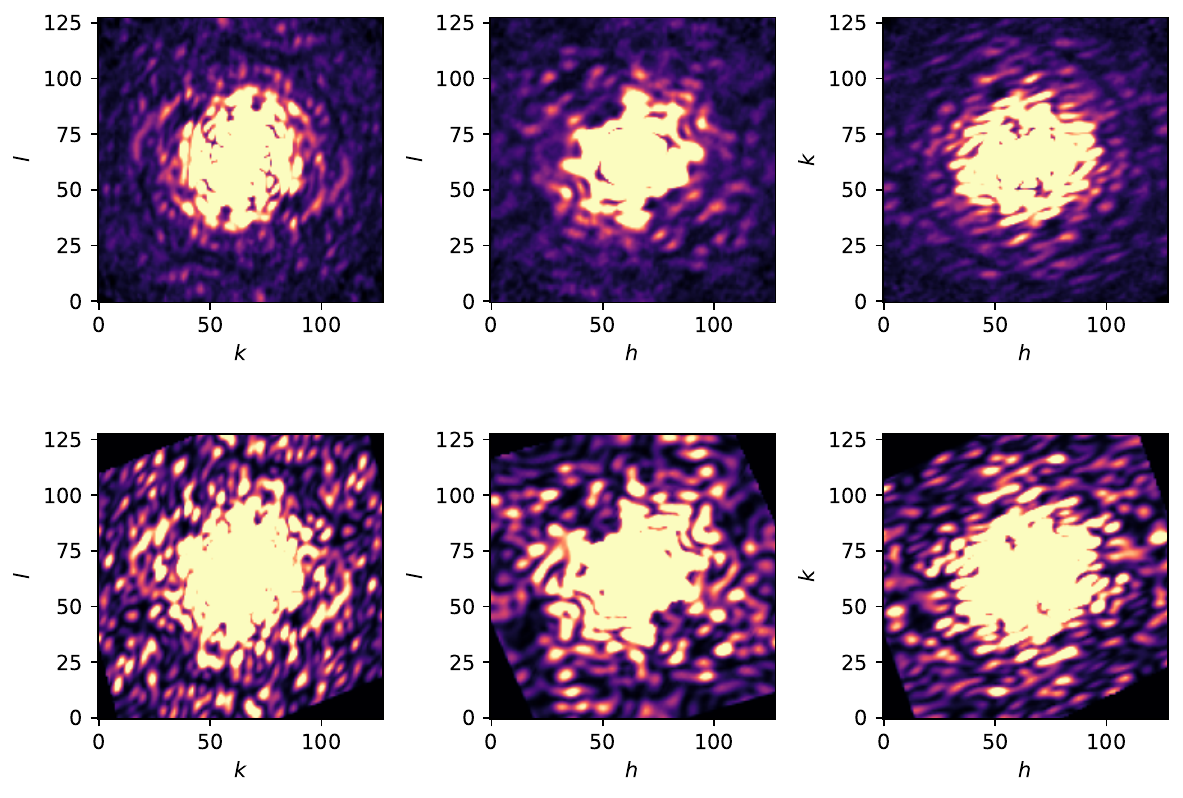}
        \label{SI_fig:fpg_100x}
    } \\
    \subfloat[Photons per pulse $10^{14}$ under the condition \texttt{FPGB}]{
        \includegraphics[width=0.6\textwidth]{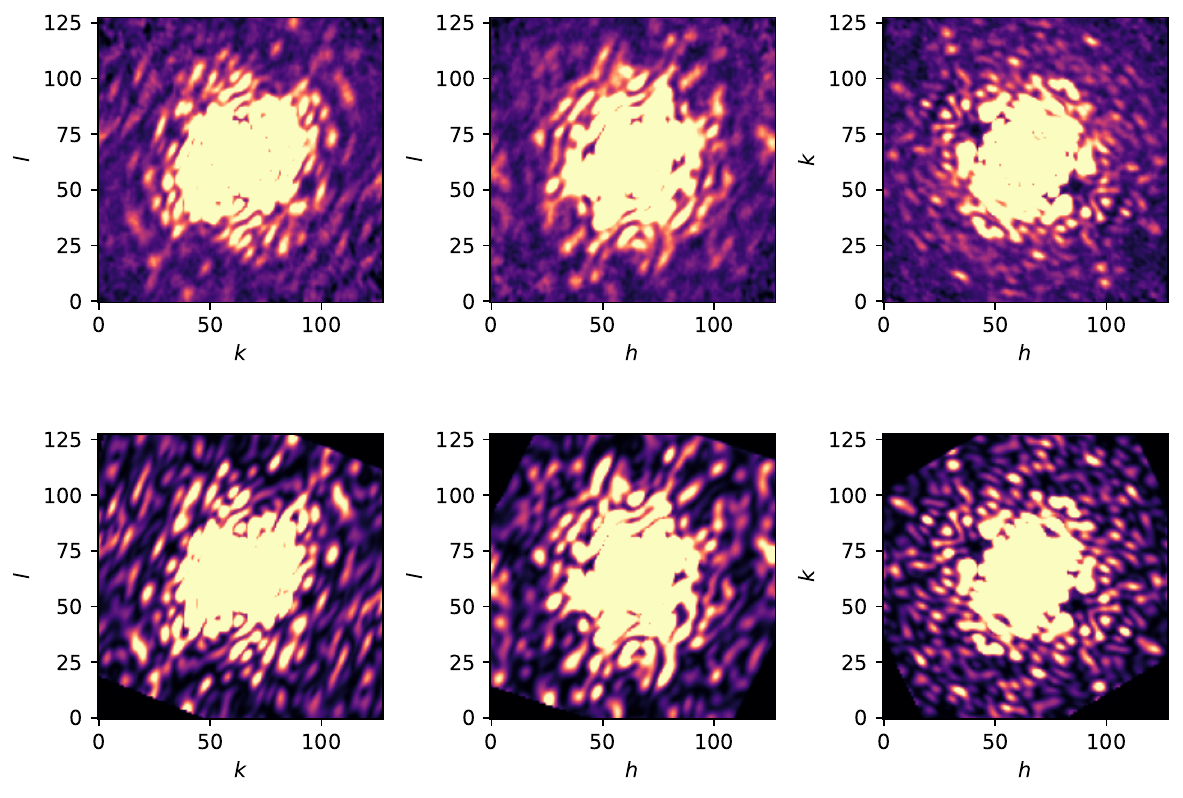}
        \label{SI_fig:fpgb_100x}
    }
    \caption{Predicted (upper panels) and true intensities (lower panels) visualized through cross-sections for models trained with an average of $10^{14}$ photons per pulse under various experimental conditions.}
    \label{SI_fig:intensity_100x}
\end{figure*}

To further investigate the impacts of various experimental conditions, we present the detailed comparisons for intensity prediction in Figure \ref{SI_fig:intensity_10x} and \ref{SI_fig:intensity_100x}. In particular, every row in all panels is displayed with an upper threshold capped at $0.2\%$ of the maximum intensity in order to reveal features at higher spatial frequencies.

We observe that the predicted intensities under lower photons per pulse come with stronger background noises when more artifacts are added into the dataset, as evidenced from panel (a) to (c) in Figure \ref{SI_fig:intensity_10x}. For the case with conditions \texttt{FPGB}, the level of background noise becomes close to physical signals at large momentum points, thus strongly affecting the phase retrieval reconstruction as shown in the main text. Meanwhile, even though the high-frequency features are not well captured in the case of conditions \texttt{FP}, the phase retrieval still manages to come up with a satisfactory reconstruction with accurately predicted lower-frequency features and without negative influences from vague and noisy high-frequency signals.

When the diffraction patterns are taken with higher photon counts, even though similar rises of background noise exist (Figure \ref{SI_fig:intensity_100x}), the physical signals are more prominent compared with the condition with lower photon counts. This is particularly obvious for the \texttt{FPGB} case as shown in panel (c), where one can tell signals from the background clearly even at the farthest $q$ points. Thus, even with locally poor predicted intensities at the radius close to the beam stop mask size, the overall clearer intensities with higher signal-to-noise ratios lead to more robust phase retrieval reconstructions.

\section{Training and validation losses}
\label{SI_sec:loss_hist}

\begin{figure*}
    \centering
    \includegraphics[width=0.9\linewidth]{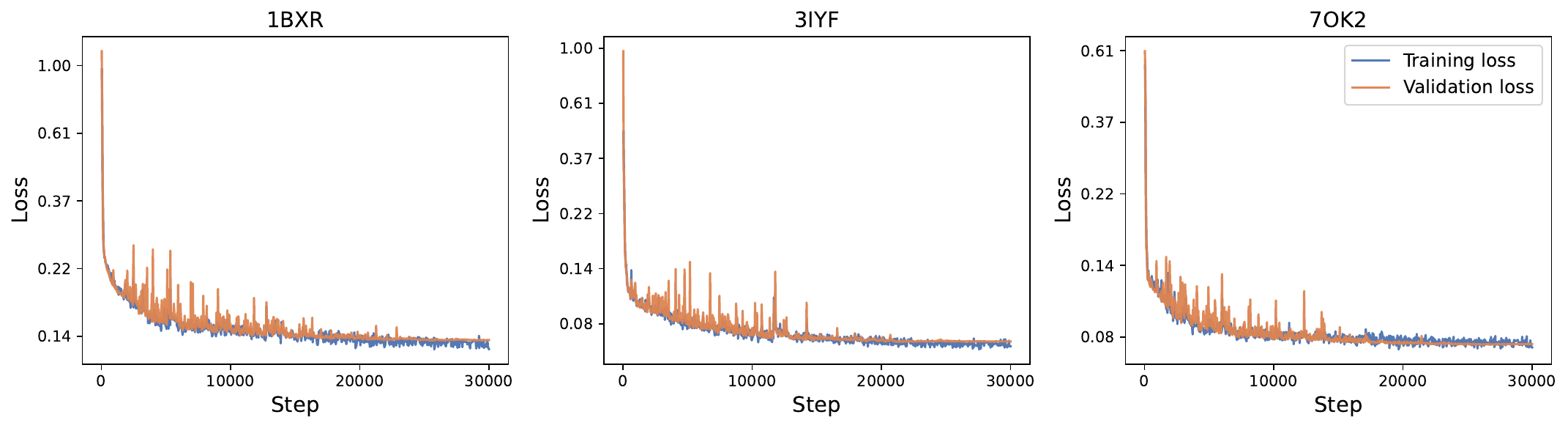}
    \caption{Training and validation loss histories for three particles under the condition \texttt{FPG} with an average of $10^{14}$ photons per pulse.}
    \label{SI_fig:loss_history}
\end{figure*}

Three representative loss histories are shown in Figure \ref{SI_fig:loss_history}, where great generalizabilities are evidenced by the close trends between training and validation loss.

\section{Uncertainties brought by phase retrieval}
\label{SI_sec:full_test_pdb_generalizability}

\begin{figure*}
    \centering
    \includegraphics[width=0.9\linewidth]{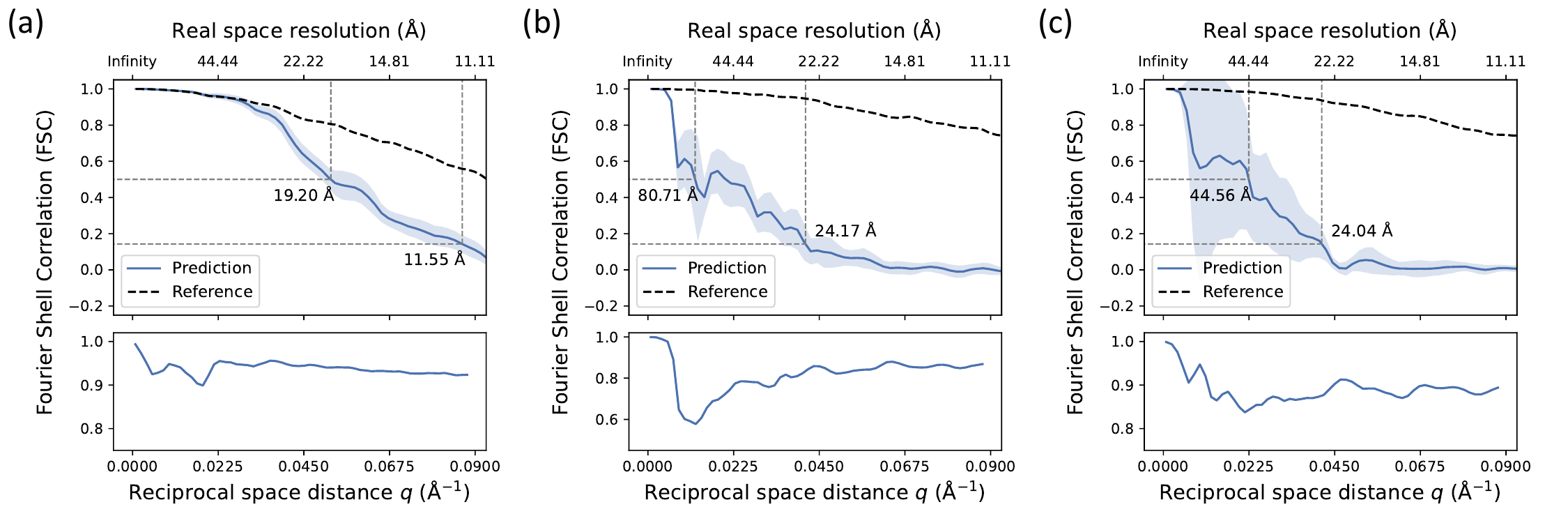}
    \caption{Model performances evaluated over 1BXR, 3IYF, and 7OK2, as displayed in panel (a), (b), and (c), respectively. The solid curves in upper panels represents the mean FSC metrics across 10 independent runs, while the shaded areas stand for $\pm 1$ standard deviations.}
    \label{SI_fig:pdb_generalizability_mean}
\end{figure*}

In Figure \ref{SI_fig:pdb_generalizability_mean}, we present the mean values and standard deviations of Fourier shell correlations (FSC) evaluated from the phase retrieval reconstructed $\rho_{\mathrm{pred}}(\mathbf{r})$ and their true conterparts $\rho_{\mathrm{true}}(\mathbf{r})$.

\section{Beam stop mask}
\label{SI_sec:beam_stop_mask}

\begin{figure*}
    \centering
    \includegraphics[width=0.4\linewidth]{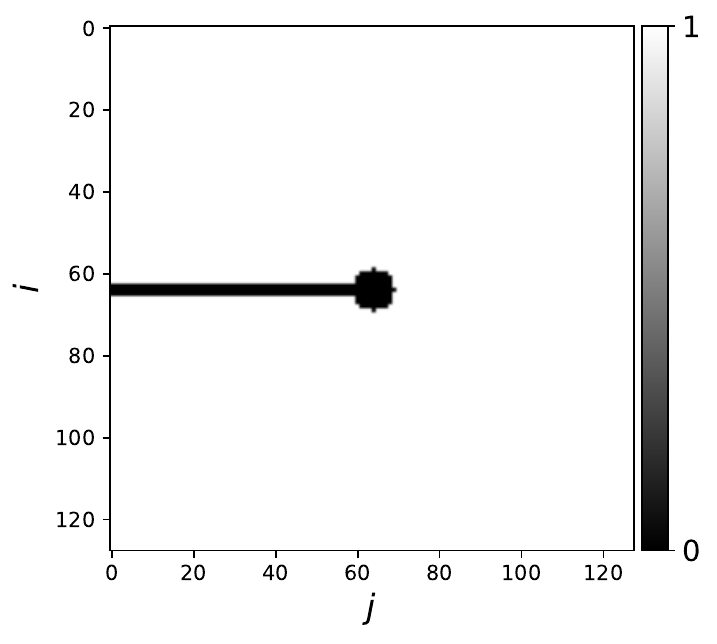}
    \caption{The binary mask used for representing the presence of a beam stop.}
    \label{SI_fig:beam_stop_mask}
\end{figure*}

We display the applied binary mask for simulating the presence of a beam stop mask in Figure \ref{SI_fig:beam_stop_mask}.

\section{FSC metrics of 3IYF and 7OK2}
\label{SI_sec:fsc_3iyf_7ok2}

\begin{figure*}
    \centering
    \includegraphics[width=0.9\linewidth]{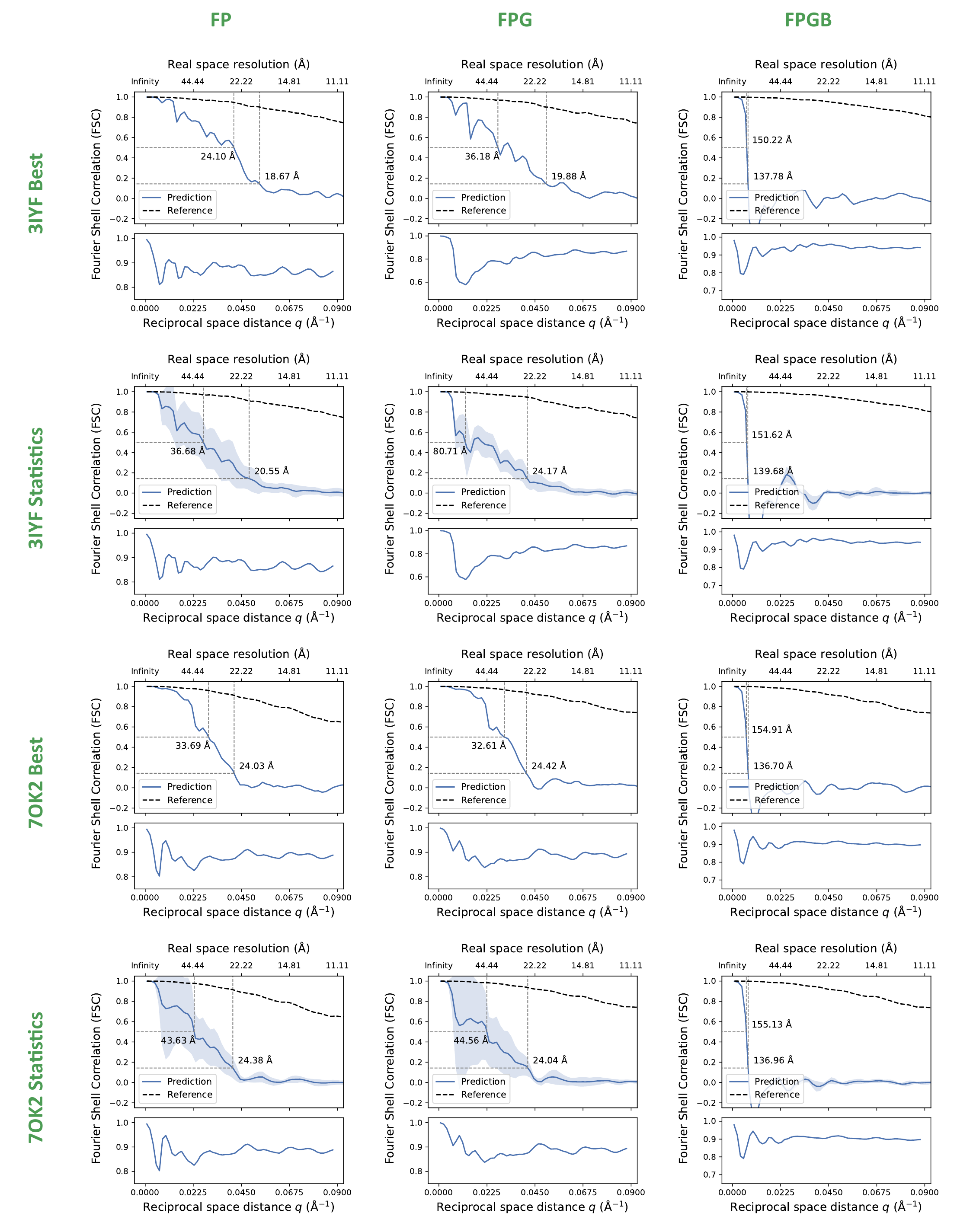}
    \caption{Summary of FSC metrics for 3IYF and 7OK2 under various experimental conditions.}
    \label{SI_fig:3iyf_7ok2_metrics}
\end{figure*}

We present the summary of FSC metrics for the 3IYF and 7OK2 in Figure \ref{SI_fig:3iyf_7ok2_metrics} to complement the results in Figure 3 in the main text. Given the more complex structures of 3IYF and 7OK2 compared to 1BXR, the model performances become more vulnerable to the most challenging experimental artifact, the beam stop mask, where no successful phase retrieval reconstruction is achieved. 

However, we note that the FSCs evaluated on predicted intensities all remain above 0.5, indicating generally good predictions on intensities and orientations. 

